\documentclass[sigplan]{acmart}

\title{}
\shorttitle{} %
\author{}
\shortauthors{} 

\definecolor{ringblue}{RGB}{220, 237, 255}

\usepackage{xcolor}
\usepackage[table]{xcolor}

\usepackage{amsmath}

\usepackage{tikz}

\definecolor{mylightblue}{RGB}{200,217,255}

\definecolor{clearblue}{RGB}{91,155,213}

\definecolor{tina_bg}{RGB}{255, 248, 232}

\begin{document}







\fcolorbox{white}{mylightblue}{

    \parbox{\dimexpr\textwidth-2\fboxsep-2\fboxrule}{
        \vspace{15pt} 
        
       
       \LARGE \color{black}\hspace{5pt}\textbf{Tiny Model, Big Logic: Diversity-Driven Optimization Elicits 
       Large-Model \\ 
       \hspace*{5cm}Reasoning Ability in \color{red}\textbf{VibeThinker-1.5B}}
        
        \vspace{5pt} 
        
        \color{black}
        
        \begin{center}
             \normalsize Sen Xu, Yi Zhou, Wei Wang, Jixin Min, Zhibin Yin, \\Yingwei Dai, Shixi Liu, Lianyu Pang, Yirong Chen, Junlin Zhang \\
             \vspace{8pt}
              Sina Weibo Inc.
            \end{center}

        \vspace{8pt} 
        
       \small OpenAI's o1 model has established a new reasoning paradigm through Long Chain-of-Thought, marking significant progress in reasoning technology. The prevailing approach continues to rely on scaling model parameters to enhance capabilities—for example, DeepSeek R1 reaches 671B parameters, and Kimi k2 exceeds 1T. A mainstream consensus holds that small models inherently lack robust reasoning capabilities. This technical report challenges that notion, demonstrating that this assumption may be incorrect. We introduce VibeThinker-1.5B, a 1.5B-parameter dense model developed using an innovative post-training methodology centered on the “Spectrum-to-Signal Principle (SSP)”. This framework systematically enhances output diversity by first employing a “Two-Stage Diversity-Exploring Distillation” in the SFT phase to generate a broad spectrum of solutions, followed by the “MaxEnt-Guided Policy Optimization (MGPO)” framework in the RL phase to amplify the correct signal. With a total training cost of \$7,800, VibeThinker-1.5B demonstrates superior reasoning capabilities compared to closed-source models Magistral Medium and Claude Opus 4, while achieving performance on par with open-source models like GPT OSS-20B Medium. Most remarkably, it surpasses the initial DeepSeek R1 model—which is over 400 times larger—across three challenging mathematical benchmarks: AIME24 (80.3 vs. 79.8), AIME25 (74.4 vs. 70.0), and HMMT25 (50.4 vs. 41.7). This marks a substantial improvement over its base model, which scored 6.7 on AIME24, 4.3 on AIME25, and 0.6 on HMMT25. Similarly, on the LiveCodeBench V6 coding benchmark, VibeThinker-1.5B achieves a score of 51.1, slightly outperforming Magistral Medium's 50.3 and a substantial improvement over the base model's 0.0. These findings demonstrate that small models can achieve reasoning capabilities comparable to large models, drastically reducing the associated costs for training and inference and thereby democratizing access to advanced AI research and accelerating technological progress. We release our post-trained model checkpoint to support future research.   
       
        \vspace{10pt} 
        
        \noindent \textbf{Date}:\hspace{0.1cm}Nov. 7, 2025 \\
        \textbf{Github}:\hspace{0.1cm}\textcolor{clearblue}{\url{ https://github.com/WeiboAI/VibeThinker}} \\
        \textbf{HuggingFace}:\hspace{0.1cm}\textcolor{clearblue}{\url{https://huggingface.co/WeiboAI/VibeThinker-1.5B} }\\ 
        \textbf{Mail}: \hspace{0.1cm}\{ xusen1, junlin6 \}@staff.weibo.com
        
        
        \vspace{10pt} 
    }
    }


\begin{figure}[ht!]
  \centering
  \includegraphics[height=9cm, width=16cm]{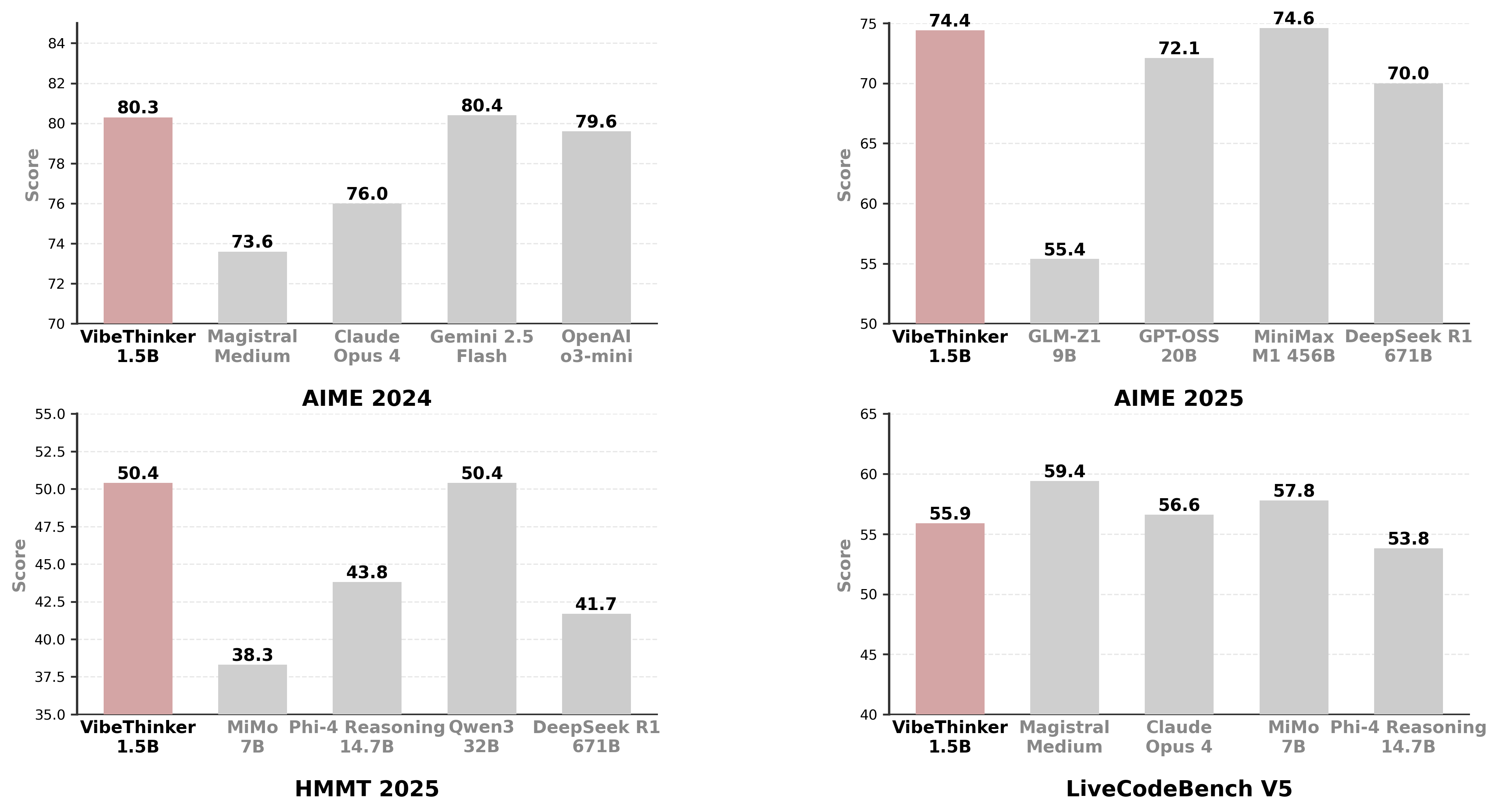}
  \caption{Performance of VibeThinker-1.5B versus competing models on representative benchmarks.}
  \label{fig:logo}
\end{figure}


 




\section{Introduction}
OpenAI o1\cite{openai2024o1} pioneered the Large Reasoning Model (LRM) paradigm, significantly enhancing the logical reasoning capabilities of large language models through reinforcement learning and extended chain-of-thought processes. It achieved human-level performance in complex domains such as mathematical theorem proving, clinical diagnosis, and competitive programming, reaching expert-level proficiency in scientific reasoning and significantly surpassing previous state-of-the-art models. Subsequent open-source projects, including DeepSeek-R1\cite{guo2025deepseek}, Qwen3\cite{yang2025qwen3}, and GLM 4.5\cite{zeng2025glm45}, further advanced these capabilities through optimized training data selection strategies and improved reinforcement learning (RL) algorithms. These efforts established both RL Scaling and test-time scaling as key optimization strategies. By allocating more computational resources during training and inference—guided by refined reward models and multi-path exploration—model accuracy on real-world problems consistently improves. The LRM paradigm has thus redefined scaling laws for reasoning-centric training, accelerating progress toward robust, generalist AI systems capable of tackling open-ended intellectual challenges.

\noindent Industry consensus holds that scaling model parameters—exemplified by DeepSeek R1 (671B)\cite{guo2025deepseek}, Kimi-K2\cite{kimi2025k2}, and Qwen3-Max\cite{alibaba2024qwen3max} (>1T)—is essential for enhancing capabilities like logical reasoning, with small models deemed significantly inferior. However, we challenge this view by investigating whether compact models (e.g., 1.5B "Tiny models") can match the reasoning performance of state-of-the-art large models. An affirmative answer would imply that the industry need not rely solely on extreme parameter scaling but could prioritize advancing small models, drastically reducing training/inference costs, energy consumption, and environmental impact. While recent small models like DeepscaleR\cite{luo2025deepscaleR}, ProRL\cite{liu2025prorl}, and Qwen3-1.7B\cite{yang2025qwen3} show promise, they have yet to fully exploit the potential of logical reasoning. This technical report demonstrates that a 1.5B model, with appropriate training, can achieve reasoning parity with today’s largest models. We open-source VibeThinker-1.5B not as a deployable solution but to prove that small models possess far greater reasoning capabilities than previously assumed.

\noindent In this report, we introduce VibeThinker-1.5B, a tiny dense language model with powerful reasoning capabilities, whose development is guided by the “Spectrum-to-Signal Principle (SSP)”. This principle redefines the post-training pipeline by decoupling the objectives of Supervised Fine-Tuning (SFT) and Reinforcement Learning (RL) into two distinct, synergistic phases. First, the SFT stage operates as the "Spectrum Phase," where we employ a “Diversity-Exploring Distillation” methodology to cultivate a broad spectrum of diverse solutions, rather than solely maximizing single-shot accuracy. Subsequently, the RL stage functions as the "Signal Phase," utilizing the “MaxEnt-Guided Policy Optimization (MGPO)” framework to identify and amplify the most effective reasoning paths from this pre-established spectrum. MGPO dynamically prioritizes training on problems where the model exhibits high uncertainty, thereby maximizing learning efficiency. By systematically integrating these two phases, our approach establishes diversity as the central technical design principle, enabling VibeThinker-1.5B to achieve robust performance that surpasses conventional training paradigms.

\begin{figure}[ht!]
  \centering
  \includegraphics[height=9cm, width=16cm]{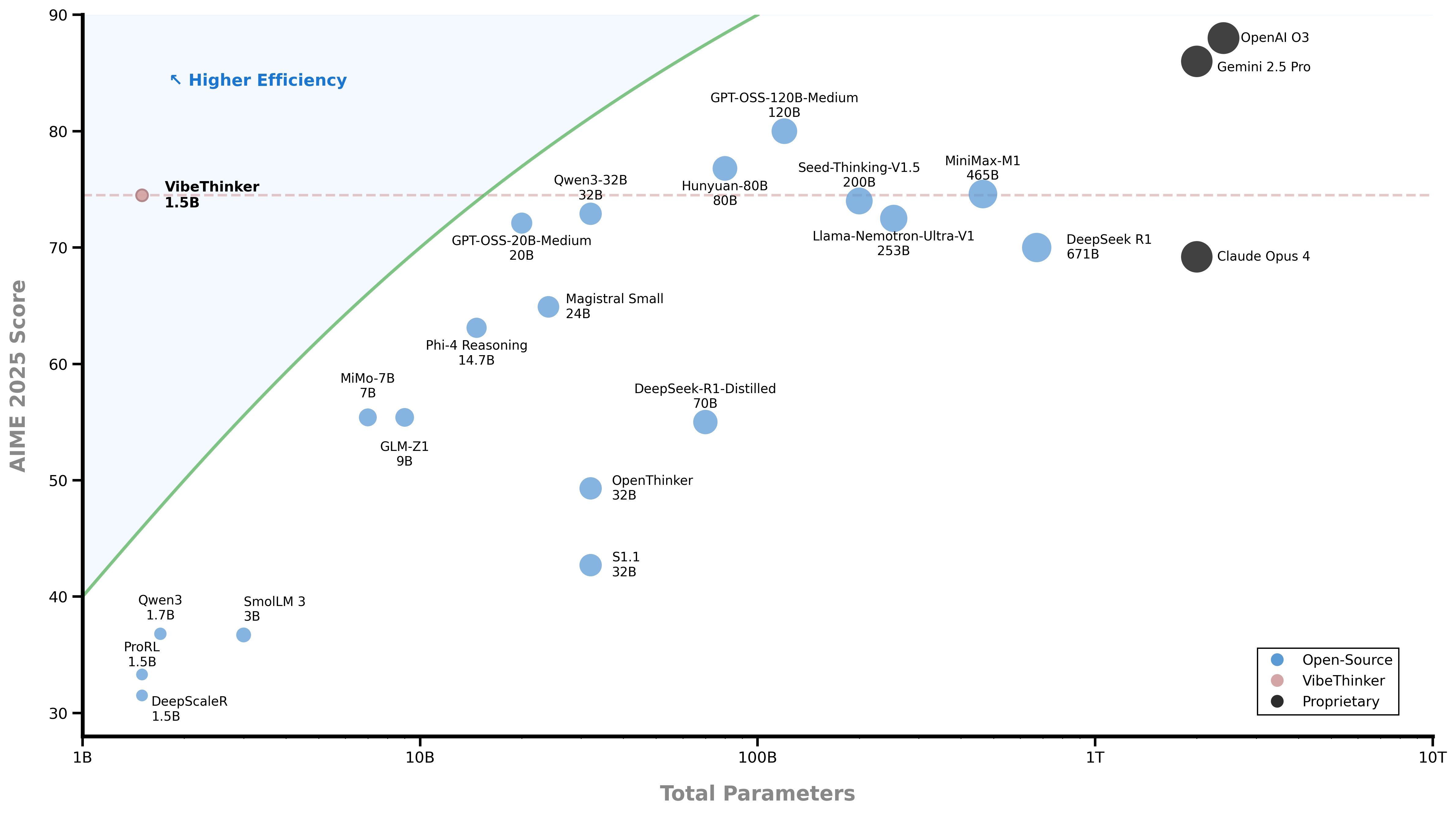}
  \caption{ VibeThinker-1.5B demonstrates remarkable efficiency, surpassing much larger and stronger models on the challenging AIME25 benchmark. It achieves a score of 74.4, outperforming strong baselines such as GPT-OSS-20B-Medium (72.1/20B), DeepSeek-R1-0120 (70.0/671B), and Seed-Thinking v1.5 (74.0/200B).}
  \label{fig:logo}
\end{figure}

\noindent VibeThinker-1.5B redefines the efficiency frontier for reasoning models, achieving state-of-the-art performance in mathematical and coding tasks with only 1.5B parameters—100× to 600× smaller than giants like Kimi K2 (1000B+) and DeepSeek R1(671B). Trained for under \$8,000, VibeThinker-1.5B, the leading model in the sub-3B category, demonstrates superior performance on challenging mathematical and coding benchmarks, surpassing many significantly larger and more powerful models. Our model exhibits exceptional performance, consistently exceeding both larger reasoning models and state-of-the-art non-reasoning models. On the demanding AIME24, AIME25, and HMMT25 mathematical tests, it surpasses the massive DeepSeek R1-0120 (over 400x larger) with scores of 80.3 vs. 79.8, 74.4 vs. 70.0, and 50.4 vs. 41.7, respectively. This competitive edge extends to non-reasoning models, where it significantly outperforms Kimi-K2-Instruct on the AIME24 math benchmark (80.3 vs. 69.6) and GPT-4.1 on the LiveCodeBench v6 coding benchmark (51.1 vs. 44.7).

\noindent Our key insight is that meticulous algorithmic design enables compact models (e.g., 1.5B parameters) to achieve logical reasoning capabilities in mathematics, code, and scientific tasks comparable to models tens to hundreds of times larger. This reveals the underestimated potential of small-scale models in reasoning.  While the study by Belcak et al. (2025) \cite{belcak2025small} proposes small models as the future of autonomous agents, these remain theoretical. We substantiate this perspective empirically by developing a 1.5B parameter "Tiny" model with strong reasoning performance.

\noindent Powerful small models not only significantly reduce the costs associated with training large models and performing online inference—thereby promoting broader adoption of AI applications—but also address a critical issue of research accessibility. The current emphasis on scaling model parameters has concentrated cutting-edge research within a handful of technology companies (e.g., OpenAI, Anthropic, Google, xAI) that possess vast computational resources. This monopolization marginalizes many corporations and universities, which often host abundant high-quality research talent but lack sufficient hardware, preventing them from contributing to the frontier of large model development. If small models demonstrate competitive performance with large models across multiple domains, the significantly lower development costs would broaden participation, enabling a wider research community to contribute and thus accelerating progress in large model technology. This would prevent core advancements from being concentrated in a few commercial entities. The advancement of compact model research therefore holds profound, albeit often implicit, significance. Although we have identified a considerable performance gap between small and large models in general knowledge benchmarks, we believe this technical challenge can be addressed through efficient methodological improvements.

\section{Preliminaries}

\textbf{Supervised Fine-Tuning (SFT).}
\hspace{0.1cm}  Supervised Fine-Tuning (SFT) adapts a pre-trained language model, parameterized by  $\theta$, to downstream tasks using a labeled dataset  $\mathcal{D} = \{(x, y)\}$ , where  $y$  denotes the reference response for a given input  $x$ . The model defines an autoregressive conditional distribution  $\pi_\theta(y|x)$ over response sequences. The training objective is to minimize the cross-entropy loss:

\[
\mathcal{L}_{\text{SFT}}(\theta) = \mathbb{E}_{(x, y) \sim \mathcal{D}} \left[ -\log \pi_\theta(y | x) \right]
\]

which is equivalent to maximizing the likelihood of the target responses under the model distribution. This process enhances task-specific alignment while preserving pre-trained knowledge.

\noindent\textbf{Group Relative Policy Optimization (GRPO).}
\hspace{0.1cm}
\noindent 
Group Relative Policy Optimization (GRPO) \cite{shao2024deepseekmath} is a reinforcement learning algorithm that extends Proximal Policy Optimization (PPO) \cite{schulman2017proximal} by replacing the critic-based advantage estimation with a group-relative mechanism. For a given query $q$, a group of $G$ responses $\{y_i\}_{i=1}^G$ is sampled from the old policy $\pi_{\theta_{\text{old}}}(\cdot|q)$. Each response $y_i$ is assigned a reward $r_i = r(q, y_i)$. The advantage for each token $t$ in response $y_i$ is then computed relative to the group's reward statistics:
\[
\mathcal{A}_{i,t}(q) = \frac{r_i - \mu_{\mathcal{G}}}{\sigma_{\mathcal{G}}}
\]
where $\mu_{\mathcal{G}}$ and $\sigma_{\mathcal{G}}$ are the mean and standard deviation of the rewards within the group. This approach reduces variance and eliminates the need for an external critic model. The optimization objective is formulated as a clipped surrogate loss, averaged over tokens and responses within the group:
\[
\mathcal{J}_{\text{GRPO}}(\theta) = \mathbb{E}_{(q,y)\sim\mathcal{D}} \left[ \mathbb{E}_{\{y_i\}_{i=1}^G\sim\pi_{\theta_{\text{old}}}(\cdot|q)}\left[\frac{1}{G}\sum_{i=1}^G\frac{1}{|y_i|}\sum_{t=1}^{|y_i|}\left(\min\left(r_{i,t}(\theta)\mathcal{A}_{i,t}(q),\text{clip}\left(r_{i,t}(\theta),1-\varepsilon,1+\varepsilon\right)\mathcal{A}_{i,t}(q)\right)\right)\right] \right]
\]
where $r_{i,t}(\theta) = \frac{\pi_\theta(y_{i,t}|q, y_{i,<t})}{\pi_{\theta_{\text{old}}}(y_{i,t}|q, y_{i,<t})}$ is the token-level probability ratio, and $\varepsilon$ controls the clipping range. For stability, a KL-divergence penalty relative to a reference policy is often added as a regularizer.

\noindent \textbf{The Relationship Between Pass@K and Diversity.}
\hspace{0.1cm} Output diversity in large language models (LLMs) refers to the variability in generated responses for a given input, which is crucial for enhancing the model's problem-solving robustness and creativity. High diversity enables the exploration of multiple reasoning paths and perspectives, preventing the model from overfitting to narrow solution patterns and increasing the likelihood of discovering novel and effective solutions. In contrast, low diversity often leads to repetitive or suboptimal outputs, limiting the model's ability to handle complex tasks requiring adaptive reasoning. This capability is particularly vital in domains like mathematical problem-solving or code generation, where exploring alternative approaches significantly improves performance.

\noindent Current research\cite{dalal2025leveraging} commonly adopts the Pass@K metric as a key indicator for assessing the diversity of outputs generated by large language models (LLMs). Pass@K measures the probability that at least one of k independently generated solutions passes a verification test (e.g., functional correctness for code or factual accuracy). Formally, for a problem  $x$  and model  $\pi_\theta$ , it is defined as:
\[
\operatorname{Pass@K} = \mathbb{E}_{x \sim \mathcal{D}, \{y_i\}_{i=1}^k \sim \pi_\theta(\cdot|x)} \left[ \max \{ R(x, y_1), \ldots, R(x, y_k) \} \right]
\]

where  $R(x, y)$ is a binary reward function indicating correctness. Diversity directly contributes to higher PASS@K scores, as a varied set of solutions broadens the exploration space and reduces the risk of all outputs failing. Empirical studies show a strong positive correlation between solution diversity and potential performance gains after reinforcement learning (RL) training, underscoring that diversity enhances the model's capacity to achieve verifiable success. Consequently, optimizing for Pass@K during supervised fine-tuning (SFT) is critical, as it encourages the development of a diverse solution repertoire, thereby improving both exploration and eventual task performance.

\section{Methodology}

\subsection{The Spectrum-to-Signal Principle for SFT-RL Synergy}
\noindent The sequential training pipeline of Supervised Fine-Tuning (SFT) followed by Reinforcement Learning (RL) is a cornerstone of modern large language model development. However, the optimal interface between these two stages remains a critically under-explored area. The prevailing, yet implicit, assumption is to select the SFT checkpoint that maximizes single-shot accuracy (Pass@1) and then use RL to further refine this same metric. We posit that this approach is suboptimal as it artificially constrains the potential performance ceiling for the subsequent RL phase.

\noindent To address this, we introduce the "Spectrum-to-Signal Principle (SSP)", a theoretical framework that redefines the roles of and the synergy between SFT and RL. In SSP, the two stages are assigned distinct, complementary objectives:

-  The Spectrum Phase (SFT): The primary goal of SFT is not to converge on a single optimal answer, but to generate a rich and diverse "spectrum" of plausible solutions. This phase maximizes the model's Pass@K metric, effectively creating a broad "candidate space" of correct answers. A model with high Pass@K provides a richer foundation for exploration, thereby raising the upper bound of what RL can achieve.

-  The Signal Phase (RL): The role of RL is then to identify and amplify the correct "signal" from within this pre-established spectrum. By receiving reward signals, the RL phase learns to increase the generation probability of the most correct and effective answers from the diverse pool provided by the SFT phase.

\noindent This principle posits that an SFT checkpoint optimized for diversity (Pass@K) is a superior prerequisite for RL, as it presents the RL algorithm with a more fertile ground for optimization compared to a narrow, Pass@1-optimized model.

\subsection{Training Pipeline}

\begin{figure}[ht!]
  \centering
  \includegraphics[width=0.9\textwidth]{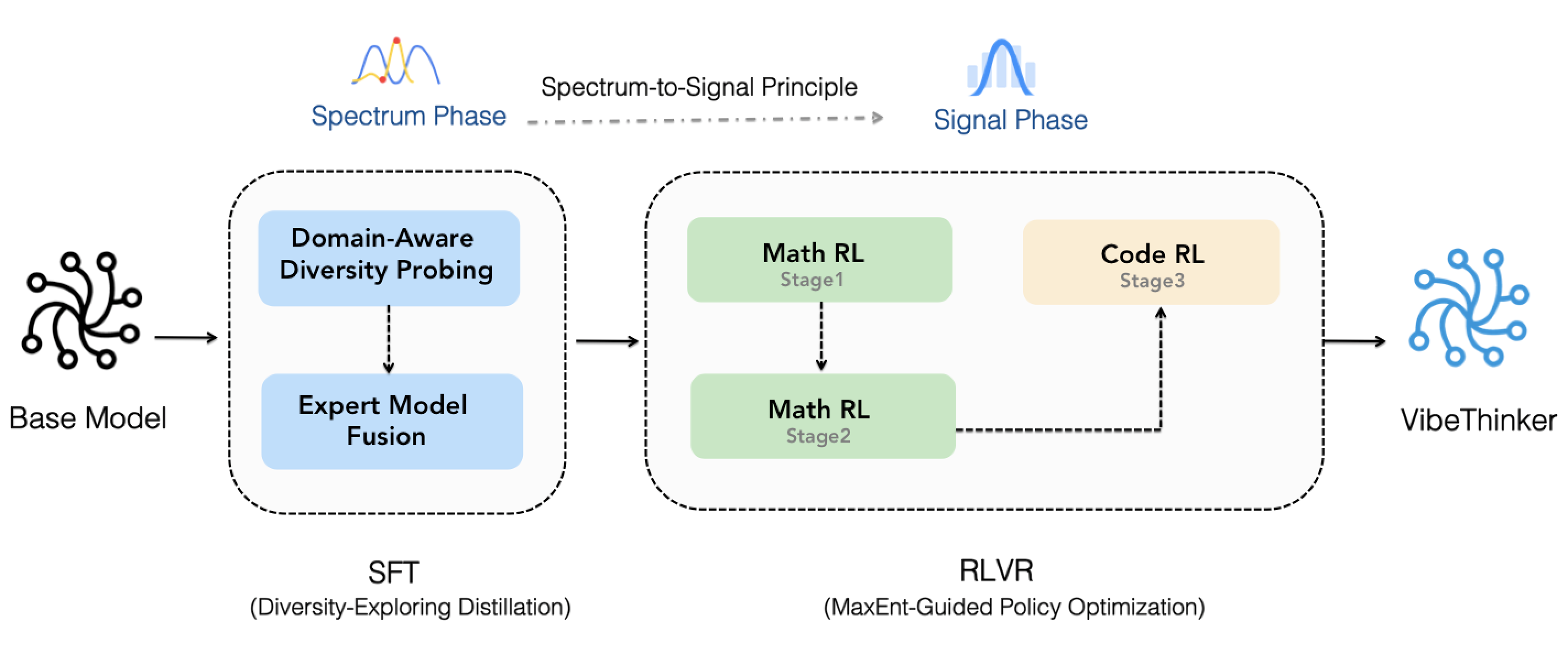}
  \caption{The Training Pipeline of VibeThinker-1.5B}
  \label{fig:logo}
\end{figure}

\noindent  Our post-training process is guided by the "Spectrum-to-Signal Principle" (SSP), which conceptualizes Supervised Fine-Tuning (SFT) and Reinforcement Learning (RL) as a sequential two-stage optimization. Unlike conventional approaches that prioritize Pass@1 in SFT, the initial stage is dedicated to cultivating a broad "spectrum" of diverse solutions. The subsequent RL phase then amplifies the correct "signal" from this pool. By intentionally fostering diversity first, this method provides a richer foundation for RL, leading to more robust reasoning and enhanced problem-solving capabilities.

\noindent The SFT stage, designated as the "Spectrum Phase", implements this principle through a "Two-Stage Diversity-Exploring Distillation" methodology. Initially, "Domain-Aware Diversity Probing" is conducted to analyze broad domains (e.g., mathematics, code) and identify sub-domains. This process pinpoints specialist SFT checkpoints that exhibit the highest diversity, as measured by Pass@K, within each sub-domain. Subsequently, "Expert Model Fusion" consolidates these optimal checkpoints using techniques like model merging. The result is a unified SFT model that embodies a maximized spectrum of plausible solutions, providing a fertile ground for the subsequent RL phase.

\noindent The RL stage, designated as the "Signal Phase", is guided by the "MaxEnt-Guided Policy Optimization (MGPO)" framework. MGPO leverages information-theoretic principles to dynamically prioritize the most pedagogically valuable problems for on-policy learning, specifically those where the policy's performance exhibits the highest uncertainty, representing a critical learning frontier. The RL phase is structured into sub-stages, beginning with mathematical reasoning within a 16K context window, expanding to 32K, and followed by code generation. MGPO enhances the exploration capacity of RL by modifying the advantage calculation to incentivize the increased generation probability of low-probability yet correct reasoning traces sampled during rollouts. This allows the model to effectively identify and amplify the correct signal from the diverse solution spectrum established by SFT.

\noindent By operationalizing the Spectrum-to-Signal Principle across the entire training pipeline, we successfully construct VibeThinker-1.5B, a small-scale model with powerful reasoning capabilities. This deliberate focus on diversity, from spectrum generation in SFT to signal amplification in RL, serves as the central theme and technical design principle of our optimization strategy, unlocking performance that surpasses the constraints of traditional training paradigms.

\subsection{Diversity-Exploring Distillation}

\noindent \textbf{Two-Stage Diversity-Exploring Distillation}
\noindent \hspace{0.1cm} To implement the spectrum phase of SSP, we propose a two-stage methodology: "Domain-Aware Diversity Probing" to identify specialist models, followed by "Expert Model Fusion" to synthesize a unified, diversity-maximized SFT model.

- \textbf{Domain-Aware Diversity Probing}
\noindent \hspace{0.1cm} To operationalize the spectrum phase, we first partition the mathematical knowledge space into $N$ distinct subdomains, $\mathcal{S} = \{S_1, S_2, \dots, S_N\}$. For instance, in our work on mathematical reasoning, we define $N=4$ with $\mathcal{S} = \{S_{\text{algebra}}, S_{\text{geometry}}, S_{\text{calculus}}, S_{\text{statistics}}\}$. For each subdomain $S_i$, we employ a capable LLM to automatically construct a specialized probing set, $D_i = \{(q_{ij}, a_{ij}) \mid j = 1, \dots, |D_i|\}$, where $q_{ij}$ is a problem and $a_{ij}$ its ground-truth answer.
During SFT training, we periodically evaluate intermediate model checkpoints $M_t$ (saved every $k$ steps) on each probing set $D_i$ using the Pass@K metric, yielding a score $P_i(t)$. The checkpoint that maximizes this metric for a given subdomain is selected as the specialist model:
\[
M_i^{*} = \arg\max_{t} P_i(t)
\]

This process yields a set of $N$ diversity-maximizing specialist models, $\{M_1^{*}, M_2^{*}, \dots, M_N^{*}\}$, each excelling at generating diverse solutions within its respective mathematical subdomain.

- \textbf{Expert Model Fusion}
\noindent \hspace{0.1cm} Having identified the specialist models, we synthesize them into a single, comprehensive SFT model optimized for the spectrum phase. This fused model, $\mathbf{M}_{\text{Merge}}^{\text{SFT}}$, is constructed as a weighted linear combination of the specialist model parameters:
$$
\mathbf{M}_{\text{Merge}}^{\text{SFT}} = \sum_{i=1}^{N} w_i M_i^{*}
$$
where the weights $w_i$ are non-negative and sum to unity ($\sum_{i=1}^{N} w_i = 1$) to preserve the model's parameter scale. In our implementation for VibeThinker-1.5B, we employ an unweighted averaging scheme where $w_i = \frac{1}{N}$ for all $i$, ensuring equitable integration of the diverse capabilities from all subdomains.

\noindent Empirically, our findings confirm the core tenets of the Spectrum-to-Signal Principle. The model $\mathbf{M}_{\text{Merge}}^{\text{SFT}}$, synthesized via Pass@K maximization, demonstrates a remarkable dual optimization: it attains state-of-the-art performance on both the Pass@K (diversity) and Pass@1 (accuracy) metrics. This result indicates that the optimization of a model's generative spectrum is not at the expense of its primary signal strength. On the contrary, a broader spectrum appears to reinforce the most correct pathways. This establishes a powerful paradigm where the SFT “spectrum” phase creates a synergistic foundation, maximizing the potential for performance gains in the subsequent RL “signal” phase.

\subsection{MaxEnt-Guided Policy Optimization (MGPO)}

In reinforcement learning from human feedback (RLHF), particularly for complex reasoning tasks, the selection of training data is paramount. A static dataset often presents a uniform challenge, failing to adapt to the evolving capabilities of the policy model. We propose "MaxEnt-Guided Policy Optimization (MGPO)", a novel framework that leverages information-theoretic principles to dynamically identify and prioritize the most pedagogically valuable problems for on-policy learning. Our core hypothesis is that a problem's utility for training is maximized when the policy's current performance on it exhibits the highest uncertainty, as this signifies a critical learning frontier where the model is most receptive to exploration and improvement.

\noindent\textbf{Maximum Entropy as an Ideal for Exploration.}
\noindent \hspace{0.1cm} This process induces a binary distribution over the outcomes (correct vs. incorrect) for question \(q\). Let \(p_c(q)\) be the empirical probability of a correct answer, derived from the \(G\) rollouts:

\[
p_c(q) = \frac{1}{G} \sum_{i=1}^{G} \mathbb{I}(r_i = 1)
\]
where \(\mathbb{I}(\cdot)\) is the indicator function. According to the principle of maximum entropy, this distribution is most "uninformed" or uncertain when its entropy is maximized. For a binary distribution, the maximum entropy occurs when \(p_c(q) = 0.5\). In this state, the model is completely uncertain about the correct answer; it is neither consistently correct nor consistently wrong. We argue that this state of maximum uncertainty represents a problem with optimal "exploratory value". Such a problem lies at the very edge of the model's current capabilities, making it an ideal candidate for policy optimization.

\noindent\textbf{Entropy Deviation Regularization.}
\noindent \hspace{0.1cm} While directly using the Shannon entropy \(H(q)\) is an intuitive approach, we propose a more targeted weighting scheme that explicitly measures and penalizes deviation from the ideal maximum-entropy state. We term this \textbf{"Entropy Deviation Regularization"}. The core idea is to define a "distance" from the ideal distribution and use this distance to modulate the learning signal.
We define the \textbf{"Max-Entropy Deviation Distance"}, \(D_{\text{ME}}(p_c(q) \| p_0)\), as the Kullback-Leibler (KL) divergence between the observed accuracy \(p_c(q)\) and the target maximum-entropy distribution \(p_0 = 0.5\). This metric effectively quantifies how much the model's performance has deviated from the state of optimal uncertainty. The distance is calculated as:

\[
D_{\text{ME}}(p_c(q) \| p_0) = p_c(q) \log \frac{p_c(q)}{p_0} + (1 - p_c(q)) \log \frac{1 - p_c(q)}{1 - p_0}
\]
Using this distance, we construct a weighting function, \(w_{\text{ME}}\), that assigns the highest weight to questions where the accuracy is closest to 0.5 and exponentially suppresses the weight as the accuracy moves towards 0 or 1:
\[
w_{\text{ME}}(p_c(q)) = \exp(-\lambda \cdot D_{\text{ME}}(p_c(q) \| p_0)), \quad \text{where} \ p_0 = 0.5, \lambda \geq 0
\]
Here, \(\lambda\) is a regularization coefficient that controls the sharpness of the weighting. When \(\lambda=0\), \(w_{\text{ME}}=1\) and the algorithm degrades to standard GRPO. As \(\lambda\) increases, the penalty for deviating from the 0.5 accuracy becomes more severe, focusing the training more aggressively on the most uncertain problems.
This weighting function is applied directly to the advantage term within the GRPO objective. The updated advantage for each rollout \(j\) in a group for question \(q\) is:
\[
\mathcal{A}'_j(q) = w_{\text{ME}}(p_c(q)) \cdot \mathcal{A}_j(q)
\]

\noindent\textbf{The MGPO Optimization Objective}
\noindent \hspace{0.1cm} MGPO integrates this entropy deviation weight directly into the GRPO optimization process. The modified objective function, \(\mathcal{J}_{\text{MGPO}}(\theta)\), is formulated as:
\[
\mathcal{J}_{\text{MGPO}}(\theta) = \mathbb{E}_{(q,y)\sim\mathcal{D}} \left[ \mathbb{E}_{\{y_i\}_{i=1}^G\sim\pi_{\theta_{\text{old}}}(\cdot|q)}\left[\frac{1}{G}\sum_{i=1}^G\frac{1}{|y_i|}\sum_{t=1}^{|y_i|}\left(\min\left(r_{i,t}(\theta)\mathcal{A}'_{i,t}(q),\text{clip}\left(r_{i,t}(\theta),1-\varepsilon,1+\varepsilon\right)\mathcal{A}'_{i,t}(q)\right)\right)\right] \right]
\]

\noindent In this formulation, the standard GRPO objective is modulated by the entropy deviation weight \(\mathcal{A}'_j(q)\). This creates an implicit curriculum learning mechanism where the model is automatically steered towards focusing its gradient updates on questions for which its current performance is most ambiguous. By doing so, MGPO ensures that the computational budget is spent on the most impactful learning opportunities, fostering more robust exploration and efficient policy improvement grounded in the principle of maximizing uncertainty.





\subsection{Training Data \& Decontamination}
For model training, the majority of data was derived from publicly available open-source datasets, while a minor portion originated from proprietary synthetic data generated internally to enhance domain-specific coverage and robustness.

\noindent To ensure the impartiality of model evaluation and the authenticity of generalization capabilities, we implemented rigorous data decontamination procedures on the training data during both the Supervised Fine-Tuning (SFT) and Reinforcement Learning (RL) stages. The primary objective of this process is to eliminate semantic overlap or information leakage risks between the training data and evaluation sets, thereby preventing assessment biases caused by data contamination. The specific operations include:

(1). Text Standardization and Preprocessing: Prior to matching, texts were normalized by removing irrelevant punctuation, symbols, and unifying letter cases to reduce noise interference and enhance matching accuracy.

(2). Semantic Decontamination: We employed 10-gram matching to identify and exclude training samples potentially overlapping semantically with evaluation sets. By reducing the n-gram length, we increased matching sensitivity to capture local semantic similarities more precisely, thereby strengthening the rigor of decontamination.

\noindent These measures significantly mitigate the risks of information leakage, ensuring that the performance of the model in the core evaluations, such as mathematical reasoning (for example, AIME24 / AIME25 \cite{aime2025problems}) and code generation (e.g., LiveCodeBench\cite{jain2025livecodebench})—faithfully reflects its true generalization and reasoning capabilities.

\noindent There is an ongoing debate regarding whether certain base models have undergone adequate data decontamination. Some study by Wu et al. (2025)\cite{wu2025reasoning}  attempted to demonstrate that the Qwen 2.5-7B\cite{yang2024qwen2math} model suffers from MATH500\cite{lightman2024verify} data leakage, arguing that this could explain why even incorrect RL reward signals might lead to seemingly good results. In contrast, another study by Wu et al. (2025)\cite{wu2025model}  contends that data leakage is not the primary factor; instead, it emphasizes the critical role of model-task alignment—defined as the congruence between a model's inherent capabilities and the requirements of a task. According to this view, strong innate model abilities can be effectively activated with minimal or even noisy training signals, especially within the model's domain of competence.

\noindent The conclusions of Study\cite{wu2025model} are more consistent with our findings. Even assuming the validity of the data leakage concerns raised in Study\cite{wu2025reasoning}, they do not adequately account for the advanced logical reasoning abilities exhibited by our VibeThinker-1.5B model. Our model is built upon Qwen2.5-Math-1.5B\cite{yang2024qwen2math}, a base model released in September 2024. Despite this, VibeThinker-1.5B demonstrated strong performance on multiple benchmarks released in 2025, including scores of 74.4 on the AIME25\cite{aime2025problems} (surpassing DeepSeek R1's 70.0) and 50.4 on the HMMT25\cite{balunovic2025matharena} (outperforming DeepSeek R1's 41.7). This performance strongly indicates that the results are not a product of training data contamination, as both the AIME25 and HMMT25 benchmarks were not publicly released until 2025. This timeline precludes their inclusion in the training data of any model finalized prior to that date, including our base mode.

\noindent Furthermore, the base model itself showed very weak capabilities in hard coding tasks, scoring 0.0 on both LiveCodeBench v5 and v6\cite{jain2025livecodebench}. Through our innovative post-training methodology, we significantly improved these scores to 55.9 on LiveCodeBench v5 and 51.1 on v6—the latter even surpassing Magistral medium\cite{mistral2024magistral}'s score of 50.3 on LiveCodeBench v6. These marked improvements across multiple new benchmarks and domains reinforce that factors beyond base model data contamination, such as targeted alignment and enhanced training techniques, are driving the performance gains.

\subsection{Training Cost}

\begin{table}[ht]
\caption{Comparison of Post-Training Costs}
\centering
\begin{tabular}{c|c|c|c|c|c}
\hline
\textbf{Models} & \textbf{Size} & \textbf{AIME25 Score}& \textbf{GPU Type} & \textbf{GPU Hours} & \textbf{Total Cost} \\
\hline
DeepScaleR & 1.5B & 31.5 & A100 & 3.8K & \$4.5K \\
MiniMax-M1 & 456B & 74.6 & H800 & 258K & \$535K \\
DeepSeek-R1 & 671B & 70.0 & H800 & 147K & \$294K \\

\hline

\textbf{\textcolor{clearblue}{VibeThinker}} & \textbf{\textcolor{clearblue}{1.5B}} & \textbf{\textcolor{clearblue}{74.4}} & H800 & 3.9K & \textbf{\textcolor{clearblue}{\$7.8K}} \\

\hline
\end{tabular}
\end{table}

\noindent We emphasize the exceptionally low post-training cost of VibeThinker-1.5B, which is directly attributable to its compact architecture of 1.5 billion parameters. Throughout the supervised fine-tuning (SFT) and reinforcement learning with verifiable rewards (RLVR) stages, the entire training process consumed approximately 3900 GPU hours on NVIDIA H800 GPUs. Based on a market rental rate of  \$2 per GPU hour for H800 instances, the total computational cost amounted to less than  \$8K.  Our 1.5B model demonstrates exceptional performance on the AIME25 benchmark, exceeding DeepSeek-R1-0120\cite{guo2025deepseek} and achieving results comparable to MiniMax-M1\cite{chen2025minimaxm1}. Although a gap to the best-in-class SOTA models persists, it has narrowed to a point where it is no longer considered an inherent limitation of small-scale models. The post-training cost of our 1.5B model is lower by one to two orders of magnitude compared to SOTA large reasoning models. While models like DeepSeek R1 and MiniMax-M1\cite{chen2025minimaxm1} incur post-training expenses of \$294K and \$535K respectively, our model's expenditure is only 1/30 to 1/60 of these figures (Table 1), demonstrating a significant breakthrough in cost-effective training.

\noindent When further considering inference service expenses, the 1.5B parameter model not only supports deployment on edge devices—such as mobile phones and vehicle-embedded systems—but also reduces inference costs by 20 to 70 times compared to state-of-the-art large-scale models. This starkly underscores the significant advantages of small models in terms of training efficiency, deployment flexibility, and overall cost-effectiveness.

\section{Evaluation and Analysis}

\begin{table*}[t]
\centering
\caption{Performance of VibeThinker-1.5B on Core Benchmarks (Small Reasoning Models)}
\label{tab:performance_no_category}
\resizebox{\linewidth}{!}{%
\begin{tabular}{l l l| cccc|cc|c} 
\toprule
\multicolumn{3}{c|}{\textbf{Model}} 
& \multicolumn{4}{c|}{\textit{Mathematics}} 
& \multicolumn{2}{c|}{\textit{Coding}} 
& \multicolumn{1}{c}{\textit{Knowledge}} \\
\cmidrule(lr){1-10}
\textbf{Name} & \textbf{Params} & \textbf{Institution} 
& \textbf{AIME 2024} & \textbf{AIME 2025} & \textbf{MATH 500} & \textbf{HMMT25} 
& \textbf{LCB v5} & \textbf{LCB v6} 
& \textbf{GPQA Diamond} \\
\midrule
L1-Max & 1.5B & CMU & 27.7 & 21.0 & 84.7 & 9.9 & -- & -- & -- \\
STILL-3 & 1.5B & RUC & 34.7 & 24.0 & 86.6 & 13.9 & -- & -- & -- \\
DeepScaleR & 1.5B & UC Berkeley & 43.1 & 31.5 & 87.8 & 19.0 & -- & 16.3 & 20.5 \\
\midrule
DeepSeek-Distill-Qwen & 1.5B & DeepSeek & 28.5 & 22.7 & 82.9 & 13.6 & 16.8 & 12.8 & 15.9 \\
FastCURL-v3 & 1.5B & Tencent & 49.6 & 34.4 & 90.5 & 21.5 & -- & 20.6 & 23.2 \\
ProRL & 1.5B & NVIDIA & 48.1 & 33.3 & 91.9 & 20.5 & 23.8 & -- & 41.8 \\
Hunyuan-0729 & 1.8B & Tencent & 56.7 & -- & 86.0 & -- & 31.5 & -- & 35.8 \\
Qwen3 & 1.7B & Alibaba & 48.3 & 36.8 & 93.4 & 23.6 & 33.2 & 26.9 & 40.1 \\
SmolLM 3 & 3B & HuggingFace & -- & 36.7 & 87.5 & 26.0 & 27.6 & 29.1 & 41.7 \\
\midrule
Base Model & 1.5B & -- & 6.7 & 4.3 & 58.5 & 0.6 & 0.0 & 0.0 & 16.4 \\
\rowcolor{blue!10}
\textbf{VibeThinker-1.5B} & 1.5B & Weibo & \textbf{80.3} & \textbf{74.4} & \textbf{95.0} & \textbf{50.4} & \textbf{55.9} & \textbf{51.1} & \textbf{46.7} \\
\bottomrule
\end{tabular}
}
\end{table*}

\subsection{ Experimental Setup}
\textbf{Benchmarks.}
\hspace{0.1cm}To comprehensively evaluate the quality of our reasoning models, we employed automated benchmarking across the following key domains:

-\textbf{Mathematics}: To assess mathematical reasoning capabilities, we employ a suite of challenging mathematical benchmarks, including MATH-500 \cite{lightman2024verify}, HMMT 2025 \cite{balunovic2025matharena}, AIME 2024, and AIME 2025\cite{aime2025problems}. To quantify model performance on the problems, we report the average pass rate across 64 repeated sampling trials as the definitive metric.

-\textbf{Coding}: We assess general programming proficiency using LiveCodeBench V5 and V6 \cite{jain2025livecodebench}. V5 comprises 279 problems from August 2024 to February 2025. Notably, V6 has two definitions: our evaluation uses the first (131 problems, February 2025–May 2025), while some comparative models may have used the second (454 problems, August 2024–May 2025), which generally yields higher scores. For both benchmarks, the final score is the average pass rate from 8 sampled outputs.

-\textbf{Knowledge}: To quantify expertise in specialized domains and complex reasoning abilities, we utilize GPQA-Diamond\cite{rein2024gpqa}—a graduate-level benchmark comprising 198 Ph.D.-level questions across biology, physics, and chemistry.

\vspace{0.2cm}
\noindent\textbf{Baselines.}
\hspace{0.1cm}To evaluate the reasoning capabilities of VibeThinker-1.5B, we compare its performance against three distinct groups of SOTA models:

(1). The most powerful sub-3B open-source reasoning models from both academia and companies, including STILL-3\cite{min2024imitate}, L1-Max\cite{aggarwal2025l1}, DeepscaleR\cite{luo2025deepscaleR}, FastCURL\cite{song2025fastcurl}, ProRL\cite{liu2025prorl}, Qwen3-1.7B\cite{yang2025qwen3}, Hunyuan 1.8B\cite{tencent2024hunyuan} and SmolLM-3 3B\cite{smollm3b}.

(2). Advanced reasoning models featuring Long-CoT capabilities are developed by both proprietary and open-source communities. Notable proprietary models include Magistral Medium (Mistral AI)\cite{mistral2024magistral}, Claude Opus 4 thinking (Anthropic)\cite{anthropic2024claude}, Gemini 2.5 Flash thinking (Google)\cite{google2024gemini}, and OpenAI o3-mini-Medium\cite{openai2024introducing}. In the open-source domain, key examples comprise MiMo-7B-RL (Xiaomi)\cite{xiaomi2025mimo}, Phi-4 Reasoning (14.7B) (Microsoft)\cite{abdin2025phi4}, Qwen3 32B thinking (Alibaba)\cite{yang2025qwen3}, DeepSeek R1 (671B)\cite{guo2025deepseek}, GPT-OSS-20B-Medium\cite{openai2024gptoss}, and MiniMax-M1\cite{chen2025minimaxm1}, among others.

(3). Top-Tier non-reasoning models, comprising the largest and highest-performing open-source and commercial models available, such as Deepseek-V3-0324, Qwen3-235B-A22B\cite{yang2025qwen3}, Kimi k2 Instruct\cite{kimi2025k2}, Claude Opus 4\cite{anthropic2024claude}, GPT 4.1\cite{openai2024gpt4}, and Gemini 2.5 Flash\cite{google2024gemini}.

\noindent This structured comparison ensures a comprehensive assessment of VibeThinker-1.5B’s reasoning performance across diverse model categories and scales.

\vspace{0.2cm}

\noindent\textbf{Evaluation Settings.}
\hspace{0.1cm}\noindent We use vLLM\cite{kwon2023efficient} as the inference backend, with a sampling temperature of 0.6 in our code (the mathematical temperature is 1.0), nucleus sampling \cite{holtzman2020curious} with top\_p = 0.95, and a maximum response length of 40k tokens. For mathematical reasoning, code generation, and domain-specific knowledge tasks, we compute Pass@1 estimates from 64, 8, and 16 samples per benchmark prompt, respectively, using strictly binary rewards. Metrics for other open-source models are sourced from evaluation results reported in their original publications or cited literature.

\subsection{Evaluation Results}

\textbf{Comparison with Small Reasoning Models.}
\noindent \hspace{0.1cm}We first compare VibeThinker-1.5B against a selection of the most powerful sub-3B reasoning models from both academia and industry. As shown in Table 2, models developed by tech companies generally outperform those from academic open-source efforts. Among 1.5B-scale models, FastCURL\cite{song2025fastcurl} and ProRL\cite{liu2025prorl} deliver the strongest results, while larger models such as Qwen3-1.7B\cite{yang2025qwen3} and Hunyuan 1.8B\cite{tencent2024hunyuan} exhibit further improvements, particularly in coding tasks—highlighting the resource disparities between academic and industrial settings.

\noindent VibeThinker-1.5B significantly outperforms its base model, Qwen2.5-Math-1.5B\cite{yang2024qwen2math}, across diverse reasoning domains. In mathematics, its AIME25 score increases from 4.3 to 74.4, while its HMMT25 score surges from 0.6 to 50.4. In coding, it achieves 55.9 on LiveCodeBench V5, up from a baseline of 0. Critically, it also demonstrates a substantial improvement in professional knowledge, with its GPQA score climbing from 16.4 to 46.7, highlighting the model's versatile potential.

\noindent VibeThinker-1.5B also redefines the performance frontier for small-scale models, achieving results that surpass not only its peers but also larger models. It more than doubles the score of the 3B SmolLM on the AIME25 benchmark (74.4 vs. 36.7) and maintains similar large margins on HMMT25 (50.4 vs. 26.0) and LiveCodeBench V5 (55.1 vs. 27.6). Against its closest peer, Qwen3-1.7B, VibeThinker-1.5B also demonstrates a substantial advantage on AIME25 (74.4 vs. 36.8) and LiveCodeBench V6 (51.1 vs. 26.9), solidifying its position as the most capable model under 3B parameters.

\noindent It is important to note that all compared models represent the top tier of reasoning capabilities. Numerous academic efforts fine-tuning Qwen2.5-Math-1.5B\cite{yang2024qwen2math} report AIME24 scores significantly below 20—for instance, the Dynamic Fine-Tuning (DFT) method\cite{wu2025generalization}, which attracted significant attention for its theoretical innovation, achieved only 6.87 on AIME24. This is not an isolated case; many improved variants of this architecture similarly struggle to surpass single-digit scores on challenging benchmarks like AIME24. Furthermore, the consistently low AIME24 scores of these Qwen2.5-Math-1.5B-based models suggest an absence of data contamination in the original Qwen2.5-Math-1.5B model, as its knowledge does not appear to overlap with such high-difficulty test items.

\begin{table*}[t]
\centering
\caption{Performance of VibeThinker-1.5B on Core Benchmarks (Large Reasoning Models)}
\label{tab:transposed_large_performance}
\resizebox{\linewidth}{!}{%
\begin{tabular}{l l l| ccc|cc|c} 
\toprule
\multicolumn{3}{c|}{\textbf{Model}} 
& \multicolumn{3}{c|}{\textit{Mathematics}} 
& \multicolumn{2}{c|}{\textit{Coding}} 
& \multicolumn{1}{c}{\textit{Knowledge}} \\
\cmidrule(lr){1-9}
\textbf{Name} & \textbf{Params} & \textbf{Institution} 
& \textbf{AIME 2024} & \textbf{AIME 2025} & \textbf{HMMT 2025} 
& \textbf{LCB V5} & \textbf{LCB V6} 
& \textbf{GPQA Diamond} \\
\midrule
Magistral-Medium-2506 & N/A & Mistral.AI & 73.6 & 64.9 & -- & 59.4 & 50.3 & 70.8 \\
Claude Opus 4 & N/A & Anthropic & 76.0 & 69.2 & -- & 56.6 & -- & 79.6 \\
Gemini 2.5 Flash & N/A & Google & 80.4 & 72.0 & -- & 61.4 & -- & 82.8 \\
OpenAI o3-mini-Medium & N/A & OpenAI & 79.6 & 74.8 & 53.0 & 66.3 & -- & 76.8 \\
\midrule
MiMo-7B-RL & 7B & XiaoMi & 68.2 & 55.4 & 38.3 & 57.8 & 49.3 & 54.4 \\
Skywork-OR1-7B & 7B & KunLun & 70.2 & 54.6 & 32.0 & 47.6 & 42.7 & -- \\
AceReason-Nemotron-1.1 & 7B & NVIDIA & 72.6 & 64.8 & 42.9 & 57.2 & 52.1 & -- \\
GLM-Z1 & 9B & ZhiPu.AI & 75.6 & 55.4 & -- & 49.1 & 42.3 & -- \\
Phi-4 Reasoning & 14.7B & Microsoft & 74.6 & 63.1 & 43.8 & 53.8 & -- & 67.1 \\
Ring-Lite & 16.8B & Ant & 76.6 & 69.1 & -- & 60.7 & -- & 61.1 \\
GPT-OSS-20B-Medium & 20B & OpenAI & 80.0 & 72.1 & -- & -- & 54.9 & 66.0 \\
Magistral-Small-2506 & 24B & Mistral.AI & 70.7 & 62.8 & 43.5 & 55.8 & 47.4 & 68.2 \\
Qwen3-32B & 32B & Alibaba & 81.4 & 72.9 & 50.4 & 65.7 & 60.1 & 68.4 \\
Llama-Nemotron-Super v1 & 49B & NVIDIA & 67.5 & 60.0 & -- & 45.5 & -- & 66.7 \\
DeepSeek-R1-Distill-Llama & 70B & DeepSeek & 70.0 & -- & -- & 57.5 & -- & 65.2 \\
Seed-Thinking v1.5 & 200B & ByteDance & 86.7 & 74.0 & -- & 64.9 & -- & 77.3 \\
Llama-Nemotron-Ultra v1 & 253B & NVIDIA & 80.8 & 72.5 & -- & 66.3 & -- & 76.0 \\
MiniMax-M1 & 456B & MiniMax & 83.3 & 74.6 & -- & 62.3 & -- & 69.2 \\
DeepSeek R1-0120 & 671B & DeepSeek & 79.8 & 70.0 & 41.7 & 65.9 & -- & 71.5 \\
\midrule
\rowcolor{blue!10}
\textbf{VibeThinker-1.5B} & \textbf{1.5B} & \textbf{Weibo} & \textbf{80.3} & \textbf{74.4} & \textbf{50.4} & \textbf{55.9} & \textbf{51.1} & \textbf{46.7} \\
\bottomrule
\end{tabular}
}
\end{table*}

\noindent \textbf{Comparison with Large Reasoning Models.}
\hspace{0.1cm} We compare VibeThinker-1.5B against several state-of-the-art reasoning models, including open-source counterparts such as Phi-4 Reasoning (14.7B), GPT-OSS-20B-Medium, MiniMax-M1, and DeepSeek R1, as well as proprietary models like Claude Opus (thinking) and OpenAI O3-mini-Medium. Despite the substantial parameter gap—ranging from 10 to hundreds of times larger than VibeThinker-1.5B—this comparison is highly illustrative. It serves to demonstrate how a meticulously designed small-scale model can challenge the conventional wisdom that performance in logical reasoning is dictated primarily by model size.

\noindent The results in Table 3 demonstrate that VibeThinker-1.5B achieves competitive performance on complex mathematical benchmarks, rivaling models with significantly larger parameter counts. Compared to proprietary models, its scores are comparable to those of O3-mini-Medium and Gemini 2.5 Flash and exceed those of Magistral Medium and Claude Opus 4 on the AIME24 and AIME25 benchmarks. When evaluated against open-source models, VibeThinker-1.5B shows consistent superiority, surpassing DeepSeek R1-0120 across all three datasets. Its performance is also closely aligned with MiniMax-M1 and superior to other models like MiMo 7B and Phi-4 Reasoning. This evidence directly challenges the long-held belief that reasoning performance is dictated primarily by model size, underscoring the untapped potential of small-scale, expertly designed architectures.

\noindent On the challenging code generation benchmarks, VibeThinker-1.5B also demonstrates competitive performance, though the gap with larger models is slightly more pronounced compared to mathematics. It achieves a performance level comparable to Magistral Medium and Claude Opus 4. We attribute this disparity primarily to our base model, Qwen-2.5-math 1.5B, which was pre-trained predominantly on mathematical data and thus had limited exposure to code. We posit that this gap is bridgeable; by strengthening the foundational code capabilities of the base model, the performance of VibeThinker could be significantly elevated.

\noindent However, it is crucial to acknowledge that on the knowledge benchmark GPQA, a substantial performance gap of 20-40 points persists between VibeThinker-1.5B and the current leading models. This suggests that smaller parameter scales may inherently limit a model's capacity to handle broad, encyclopedic general knowledge. We therefore call upon the research community to prioritize enhancing the general knowledge capabilities of small models as a critical research direction, which is essential for accelerating their widespread adoption and real-world application.

\noindent \textbf{Comparison with Top-Tier Non-Reasoning Models.}
\hspace{0.1cm}We next compare VibeThinker-1.5B against the most powerful non-reasoning models, including open-source models like Kimi K2 Instruct\cite{kimi2025k2}, Deepseek V3-0324, and Qwen3-235B-A22B\cite{yang2025qwen3} (with parameter scales ranging from 235B to 1T), as well as commercial models such as GPT-4.1\cite{openai2024gpt4}, Claude Opus 4\cite{anthropic2024claude}, and Gemini 2.5 Flash\cite{google2024gemini}. Although comparing reasoning models with non-reasoning models may seem inherently uneven, this comparison is motivated by two key considerations: first, VibeThinker-1.5B's parameter count is merely a fraction (often 1/100th or less) of these non-reasoning models; second, these large models themselves have undergone extensive reinforcement learning training with substantial math and coding data, albeit without explicit Chain-of-Thought (CoT) reasoning processes. This juxtaposition aims to highlight the significant potential of small models in reasoning tasks.

\noindent As shown in Table 4, despite its drastically smaller size, VibeThinker-1.5B surpasses all compared non-reasoning models on challenging mathematical benchmark sets and outperforms the majority in code generation tasks. These results robustly demonstrate that small models possess far greater potential in logical reasoning than previously assumed by consensus.

\noindent  However, a considerable performance disparity persists on the general knowledge benchmark GPQA, reaffirming that smaller models still face inherent limitations in handling broad domain knowledge compared to their larger counterparts.

\begin{table*}[t]
\centering
\caption{Performance of VibeThinker-1.5B on Core Benchmarks (Top-Tier Non-Reasoning Models)}
\label{tab:transposed_performance}
\resizebox{\linewidth}{!}{%
\begin{tabular}{l l l| cc| cc| c} 
\toprule
\multicolumn{3}{c|}{\textbf{Model}} 
& \multicolumn{2}{c|}{\textit{Mathematics}} 
& \multicolumn{2}{c|}{\textit{Coding}} 
& \multicolumn{1}{c}{\textit{Knowledge}} \\
\cmidrule(lr){1-3} \cmidrule(lr){4-5} \cmidrule(lr){6-7} \cmidrule(lr){8-8}
\textbf{Name} & \textbf{Params} & \textbf{Type} 
& \textbf{AIME 2024} & \textbf{AIME 2025} 
& \textbf{LCB v5} & \textbf{LCB v6} 
& \textbf{GPQA Diamond} \\
\midrule
Kimi K2 & 1.09T & Open-Source & 69.6 & 49.5 & -- & 53.7 & 75.1 \\
Deepseek V3-0324 & 671B & Open-Source & 59.4 & 46.7 & 49.2 & 46.9 & 68.4 \\
Qwen3-235B-A22B & 235B & Open-Source & 40.1 & 24.7 & -- & 37.0 & 62.9 \\
\midrule
GPT-4.1 & N/A & Proprietary & 46.5 & 37.0 & -- & 44.7 & 71.5 \\
Claude Opus 4 & N/A & Proprietary & 48.2 & 33.9 & -- & 47.4 & 81.0 \\
Gemini 2.5 Flash & N/A & Proprietary & 61.3 & 46.6 & -- & 44.7 & 71.1 \\
\midrule
\rowcolor{blue!10}
\textbf{VibeThinker-1.5B} & 1.5B & Open-Source & \textbf{80.3} & \textbf{74.4} & \textbf{55.9} & \textbf{51.1} & \textbf{46.7} \\
\bottomrule
\end{tabular}
}
\end{table*}

\section{Conclusion}
This report introduces VibeThinker-1.5B, a compact 1.5B-parameter model that challenges the prevailing scaling paradigm by achieving state-of-the-art reasoning performance at a fraction of the cost. Developed for under \$8,000, our model surpasses DeepSeek R1 on the challenging AIME25 benchmark and outperforms other leading large models on key benchmarks. This is accomplished not through extreme parameter scaling, but via innovation in post-training that enhances output diversity during supervised fine-tuning and reinforcement learning. Our results indicate that small models can possess formidable reasoning capabilities, prompting a necessary re-evaluation of Scaling Law assumptions.

\bibliographystyle{ACM-Reference-Format}
\bibliography{ref}


\end{document}